\documentclass[review]{elsarticle}

\usepackage{hyperref}
\usepackage{amssymb}
\usepackage{color}
\usepackage{makecell}
\usepackage{multirow}

\journal{Journal of \LaTeX\ Templates}

\begin{document}

\begin{frontmatter}

\title{GID-Net: DETECTING HUMAN-OBJECT INTERACTION WITH GLOBAL AND INSTANCE DEPENDENCY}

\author[mymainaddress,mysecondaryaddress,myfootnote]{Dongming Yang}
\ead{yangdongming@pku.edu.cn}
\author[mymainaddress,mysecondaryaddress]{YueXian Zou\corref{mycorrespondingauthor}}
\ead{zouyx@pku.edu.cn}
\ead[url]{http://web.pkusz.edu.cn/adsp/}
\fntext[myfootnote]{Address: Peking University, Xili Road, Nanshan District, Shenzhen, P.R.China}
\author[mythirdaddress]{Jian Zhang}
\ead{Jian.Zhang@uts.edu.au}
\author[mymainaddress,mysecondaryaddress,myforthaddress]{Ge Li}
\ead{gli@pkusz.edu.cn}
\cortext[mycorrespondingauthor]{Corresponding author}

\address[mymainaddress]{ADSPLAB, School of ECE, Peking University, Shenzhen, China}
\address[mysecondaryaddress]{Peng Cheng Laboratory, Shenzhen, China}
\address[mythirdaddress]{School of Electrical and Data Engineering, University of Technology Sydney, Australia}
\address[myforthaddress]{National Engineering Laboratory for Video Technology - Shenzhen Division}

\begin{abstract}
Since detecting and recognizing individual human or object are not adequate to understand the visual world, learning how humans interact with surrounding objects becomes a core technology. However, convolution operations are weak in depicting visual interactions between the instances since they only build blocks that process one local neighborhood at a time. To address this problem, we learn from human perception in observing HOIs to introduce a two-stage trainable reasoning mechanism, referred to as GID block. GID block breaks through the local neighborhoods and captures long-range dependency of pixels both in global-level and instance-level from the scene to help detecting interactions between instances. Furthermore, we conduct a multi-stream network called GID-Net, which is a human-object interaction detection framework consisting of a human branch, an object branch and an interaction branch. Semantic information in global-level and local-level are efficiently reasoned and aggregated in each of the branches. We have compared our proposed GID-Net with existing state-of-the-art methods on two public benchmarks, including V-COCO and HICO-DET. The results have showed that GID-Net outperforms the existing best-performing methods on both the above two benchmarks, validating its efficacy in detecting human-object interactions.
\end{abstract}

\begin{keyword}
\texttt Human-Object Interaction\sep Long-Range Dependency \sep  Semantic Reasoning\sep Convolutional Neural Network
\end{keyword}

\end{frontmatter}


\section{Introduction}
\begin{figure}
	\centering
		\includegraphics[scale=.75]{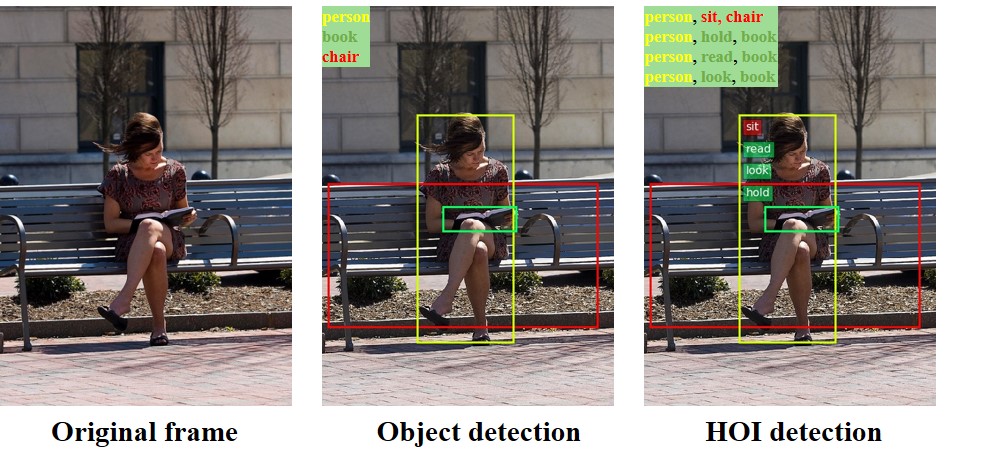}
	\caption{An example of HOI detection. Given an input frame (left) and the detected instances in the frame (middle), our method detects the interactions between each person and object (right). The individual in the frame simultaneously takes multiple interactions with surrounding objects.}
	\label{FIG:1}
\end{figure}

Recently, significant progresses have been made in detecting and recognizing individual instance in images. However, to further understand the situation in a scene, in applications like intelligent monitoring and man-machine interaction, machines need to not only detect instances but also recognize the visual relationships between them. In this paper, we tackle the task of human-object interaction (HOI) detection, which aims to infer the relationships between humans and surrounding objects. Beyond detecting and comprehending instances, e.g., object detection, action recognition and human pose estimation, learning HOIs requires a deeper semantic understanding of image contents to depict complex relationships between different human-object pairs. The task of HOI detection can be represented as detecting $\langle$ person, verb, object $\rangle$ triplets, such as $\langle$ person, eat, sandwich $\rangle$ , $\langle$ person, watch, television $\rangle$ , etc. HOI detection is related to action recognition, but presents different challenges. One of the challenge is that an individual can simultaneously take multiple interactions with surrounding objects (e.g., eating a sandwich and reading a book while sitting in a chair). Associating various objects with ever-changing roles leads to a finer-grained and diverse understanding of the current state of interaction. We show an example of the HOI detection in  Figure~\ref{FIG:1}.

Most existing approaches infer a HOI using appearance features of a person and an object extracted from convolutional neural networks (CNNs). Although convolution operations in CNNs (e.g., VGG \cite{simonyan2015very} and ResNet \cite{he2016deep}) have showed impressive ability in capturing discriminative features from different instances, they are weak in recognizing diverse relationships and fine-grained interactions between different instances. The reason is that convolution operations only build blocks that process one local neighborhood at a time, therefore, they have limited capability to learn long-range dependencies of pixels out of the local neighborhood. To alleviate the problem, recent action recognition and HOI detection algorithms exploit some additional contextual cues from the image, such as using the union of the human and object bounding boxes \cite{zhang2019large-scale, lu2016visual, zhang2017relationship}, estimating human pose and intention \cite{cheron2015p-cnn:, xu2018interact}, learning the general interactiveness knowledge \cite{li2018transferable} from multiple HOI datasets, or extracting context from the whole image \cite{mallya2016learning}. Although incorporating contextual information generally benefits feature expression, these hand-designed or untrainable contextual cues may not always be relevant for detecting interactions. For example, algorithms established on the union bounding boxes of the human and object may be not effective in identifying interactions without any overlap between the human and the object, e.g., $\langle$ person, watch, TV $\rangle$. Attending to human poses in algorithms may be not helpful for detecting HOIs with high similarity, such as $\langle$ person, drink with, cup $\rangle$ and $\langle$ person, eat with, spoon $\rangle$. Also, algorithms with additional contextual cues always bring huge computation burden. In addition to above methods, there have been some soft attention mechanisms for temporal action recognition \cite{sharma2015action, song2017an, shi2016joint} building on video settings, which are temporal related and show tiny vantage in the HOI detection task. Existing 2D based attention mechanisms for image classification \cite{jetley2018learn} or scene segmentation \cite{fu2018dual} are valuable but not suitable for detecting interactions between humans and objects.

To tackle the limitations above, we propose a trainable mechanism to infer HOIs which contains a two-stage and progressive reasoning process according to the human perception. It is noted that the human visual system is able to progressively capture long-range dependency from the scene and related instances to recognize a HOI. Considering the triplet $\langle$ person, kick, soccer ball $\rangle$ as an example, the global-level semantics from the scene (e.g., greensward) can be first captured as prior knowledge and instance-range semantics from the person and soccer ball are then learned to further recognize the verb (i.e., kick) and disambiguate other candidates (e.g., carry). Following this perception, we propose a two-stage mechanism to infer HOIs where global and instance dependency of pixels are captured progressively. Without adopting additional inputs (e.g., estimated human pose) or untrainable attention tricks, our proposal provides a powerful representation which is applicable in detecting HOIs. Aggregating these innovation synthetically, we propose GID-Net. The contributions of this work are summarized as follows:

\begin{itemize}
\item  A two-stage reasoning mechanism is proposed to learn the powerful semantic representation for detecting HOIs, referred to as GID block. We build GID block in two parts, which is the global-dependency part and instance-dependency part. The GID block is end-to-end trainable and dynamically produces reasoned features for depicting HOIs, in which both global and instance dependency of pixels are captured.
\item  An efficient framework for HOI detection is introduced  to efficiently assemble GID block, namely GID-Net. The GID-Net consists of a human branch, an object branch and an interaction branch. Semantic information in global-level and local-level are organically aggregated in each of the branches. Besides, taking ResNet-50 \cite{he2016deep} as the backbone network, GID-Net take image features from both fourth and fifth residual blocks as inputs to exploit long-range dependency. With this design, GID-Net is able to utilize the rich representation of features from different locations and different layers.
\item  By decomposing GID block into two parts (i.e., global-dependency part and instance-dependency part) and adopting different branches (i.e., human branch, object branch and interaction branch), we study the individual effect of different components in our method and provide detailed error analysis to facilitate the future research.
\end{itemize}
We validate our method on two large benchmarks, including V-COCO \cite{gupta2015visual} and HICO-DET \cite{chao2018learning}. Our method have provided obvious performance gain compared with the existing best-performing methods on these two benchmarks, achieving the state-of-the-art performance.

\section{Related work}

\subsection{Human-Object Interaction Detection}

Driven by the tremendous progress in object detection, detecting human-object interaction is going to be critical. Based on object detectors with artificial features, the earliest works exploited Bayesian models and probabilistic graphical models \cite{gupta2007objects, gupta2009observing, yao2010modeling, yao2012recognizing} for detecting HOIs. Furthermore, algorithms learned structured image representations with spatial interaction \cite{yao2010grouplet:, delaitre2011learning}, utilized action attributes and parts  \cite{yao2011human}, studied compositional models and tree-structured relations \cite{desai2012detecting}, or referred to a set of HOI exemplars  \cite{hu2013recognising}. More recently, algorithms in HOI detection have made great progress owing to the success of deep learning and the availability of large-scale HOI datasets \cite{gupta2015visual, chao2015hico:, chao2018learning}. Fill-in-the-blank questions (Visual Madlibs) model \cite{mallya2016learning} and visual translation embedding network \cite{zhang2017visual} were applied for assisting visual relationship understanding. With the assistance of zero-shot learning\cite{shen2018scaling, kato2018compositional} and graph based model \cite{qi2018learning}, structured knowledge could be incorporated into deep models to tackle HOI detection. In \cite{zhang2019large-scale}, the bounding boxes of human and object as well as their combination were fed into a multi-stream network to address visual relationship understanding. 

Very recently, several works have taken advantage of the detailed annotated and contextual cues to improve HOI detection. Auxiliary boxes \cite{gkioxari2015contextual} and human pose \cite{cheron2015p-cnn:} were employed to encode context regions from the human bounding boxes. Mallya \cite{mallya2016learning} used the whole image as contextual information and employed multiple instance learning (MIL) to predict an action label for an image. Gkioxari \cite{gkioxari2018detecting} estimated an action-specific density map to identify the locations of interacted objects based on the appearance of person. iHOI \cite{xu2018interact} modeled human pose and utilized human gaze to guide the attended contextual regions in a weakly-supervised setting, which improved the precision of localizing the interacting objects. Song \cite{song2017an} also used pose keypoints to drive attention to parts of the human body. More recently, several works have utilized the detailed human pose \cite{li2018transferable} and body parts \cite{zhou2019relation,wan2019pose} to model HOIs.Although incorporating human pose or body parts benefits feature expression, it brings huge computation burden. Specifically, Chao et al. \cite{chao2018learning} introduced a three-stream framework, exploiting the visual and spatial representations of instances and their interactions. 

As mentioned above, recent algorithms mostly extracted contextual evidence for HOI detection by employing artificial attention regions and contextual cues like auxiliary boxes, human pose, etc. Compared with these artificial attention technics, a trainable attention mechanism is a simpler and more efficient way to indicate the discriminative areas from the scene without additional inputs or huge computation burden. Inspired by the estimated density maps proposed by InteractNet \cite{gkioxari2018detecting}, our approach improves HOI detection by encoding a two-stage trainable reasoning mechanism namely GID block to automatically capture both global and instance dependency and learn the powerful semantic relationships from an image. Such a mechanism allows us to progressively infer and aggregate semantic information for detecting HOIs.

\subsection{Attention Mechanisms in HOI Detection}

There have been some works on action recognition that explored soft attention mechanism for spatio-temporal attention \cite{sharma2015action, song2017an} and temporal attention \cite{shi2016joint}. All these methods are built on video settings, using Long-Short Term Memory (LSTM) networks to predict attention maps for the current frames. However, HOI detection task mostly takes a single frame as input to predict the interactions between humans and objects, in which attention mechanisms would not rely on the temporal setting. Based on 2D feature vectors, several methods proposed trainable attention models to improve action recognition \cite{girdhar2017attentional}, object detection\cite{wang2018non-local} and image classification\cite{jetley2018learn}, respectively. Recently, some valuable efforts have been devoted to incorporate attention mechanisms into recognizing visual actions and detecting HOIs in images. Non-local operation \cite{wang2018non-local} captures long-range dependencies for action classification via computing the response at a position as a weighted sum of the features at all positions. Improved results have been shown with the designing of Box Attention mechanism \cite{kolesnikov2018detecting} and Instance-Centric Attention Network (iCAN) \cite{gao2018ican:}, demonstrating the efficacy of employing attention mechanisms in HOI detection task. Among those, iCAN \cite{gao2018ican:} delivers the state-of-the-art performance. While iCAN employed an instance-centric attention module learning to highlight informative regions using the appearance of a person or an object instance, it ignored the power of expressing HOIs via further exploiting the global semantics. On the whole, there have been no work concurrently capturing global and instance dependency of pixels and provide a progressive reasoning process in a mechanism to boost detecting interactions between humans and objects.

Analyzing of previous works shows that temporal based attention mechanisms are not suitable for detecting HOIs in static setting. Existing 2D based attention mechanisms are either without considering HOI detection or still improvable for extracting features for interactional relations. Thus, confronting with the HOI detection task, outcomes from the existing researches are limited. In this work, we propose GID-Net which contains a HOI-specific reasoning mechanism for HOI detection tasks.

\section{Proposed Method}

\subsection{GID block}

\begin{figure}
	\centering
		\includegraphics[scale=.41]{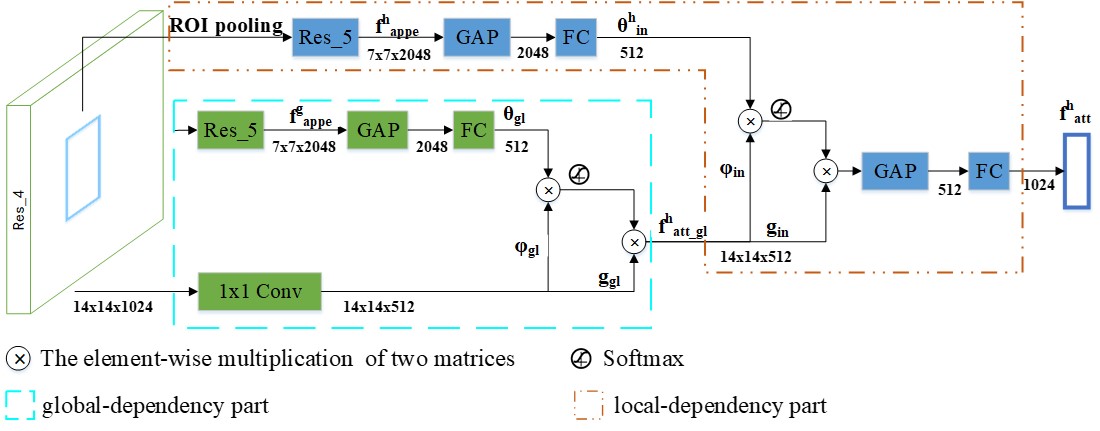}
	\caption{The proposed GID block in human branch (one of the three branches from our proposed GID-Net, detailed in section 3.2), where ResNet-50 is taken as the backbone network. GID block takes convolutional features from the image box b$_g$ and human bounding box b$_h$ as inputs and output the reasoned feature f$_{att}^h$.}
	\label{FIG:2}
\end{figure}

While recognizing a HOI, a person is able to progressively capture the long-range semantics from the scene and related instances, rather than just orderly focus repeating local neighborhoods like convolution operations do. Thus, we learn from the human perception to build a two-stage reasoning mechanism which progressively captures long-range dependency in global-level and instance-level to depict HOIs. To make the above conjecture computationally feasible, we design the GID block containing two parts to exploit both global and instance dependency. Taking human branch (one of the three branches from our proposed GID-Net, detailed in section 3.2) as an example, Figure~\ref{FIG:2} shows the detailed procedure of GID block. 

\subsubsection{Proposed structure}

ResNet-50 \cite{he2016deep} is employed as the backbone network in our implement. We divide the process of computing GID block into two parts, which are the global-dependency part and the instance-dependency part. While an interaction happens throughout the whole scene, we firstly capture global dependency from the whole scene through the global-dependency part. Afterwards, we capture instance dependency through the instance-dependency part.

\paragraph{Global-dependency part} In the global-dependency part, the shared feature extracted from the \emph{Res\_4} (the fourth residual block of ResNet-50) is used for generating the embedded image feature \emph{$\varphi_{gl}$} and g$_{gl}$ with \emph{1*1} convolutions.  \emph{$\varphi_{gl}$} and g$_{gl}$ both have \emph{512} channels, which is a half compared with that from Res\_4. This operation follows the bottleneck of ResNet \cite{he2016deep} and reduces the computation of the block by about a half. Then, the shared feature from Res\_4 passes through the Res\_5 (the fifth residual block of ResNet-50) to obtain the appearance features  f$_{appe}^g$ for the global image. Afterwords, a global average pooling (GAP) and a fully connected layer are adopted on f$_{appe}^g$, generating the embedded image feature \emph{$\theta_{gl}$}. With \emph{$\theta_{gl}$},  \emph{$\varphi_{gl}$} and g$_{gl}$ together, we adopt the matrix multiplications and softmax operation shown in Figure~\ref{FIG:2} to dynamically compute the global-level reasoned feature f$_{att\_gl}^h$.

\paragraph{Instance-dependency part} In the instance-dependency part, f$_{att\_gl}^h$ obtained from global-dependency part is employed as the embedded image feature \emph{$\varphi_{in}$} and g$_{in}$, which keeps the progressive relationship between the two reasoning prats. Meanwhile, shared feature from \emph{Res\_4} are processed with a ROI pooling according to the human bounding box and the \emph{Res\_5} block to get the feature  f$_{appe}^h$ of human appearance. A global average pooling (GAP) and a fully connected layer are then operated on  f$_{appe}^h$ to obtain the embedded human feature \emph{$\theta_{in}^h$}. Likewise, the final reasoned feature f$_{att}^h$ is dynamically generated by using the matrix multiplications and softmax operation shown in Figure~\ref{FIG:2}. With this design, it can be seen that GID block takes different convolutional features stage-by-stage which enables it to focus on different semantic information and receptive field step by step.

\subsubsection{Inference}

The goal of the matrix multiplications and softmax operation in GID block is to dynamically conduct semantic-specific reasoning and capture long-range dependency for an interaction. Following the non-local operation \cite{wang2018non-local} which is able to compute responses based on relationships between different pixels, these operations act up to the following definition:
\begin{equation}
y_i=\frac{1}{C(x)}\sum_{\forall j}f(x_i,x_j)g(x_j)
\label{equ4}
\end{equation}
Here \emph{i} is the index of an output pixel position and \emph{j} is the index that enumerates all possible pixel positions. \emph{x} and \emph{y} are the input signal (convolutional features) and the output signal, respectively. A pairwise function \emph{$f(x_i,x_j)$} computes a scalar (representing relationship such as affinity) between \emph{i} and all \emph{j}. The unary function \emph{$g(x_j)$} computes a representation of the input signal at the position \emph{j}. The response is normalized by a factor \emph{$C(x)$}. 

Differing from iCAN \cite{gao2018ican:} which highlights informative regions using the appearance of a person or an object, GID-Net expresses HOIs via exploiting long-range dependency in global-level and instance-level and ultimately provides fused semantic-specific reasoning between visual targets (i.e., scene, human, object). As shown in Figure~\ref{FIG:2}, different pixel positions are considered in the operations from global-dependency part and instance-dependency part, and progressive strategy is adopted to compute relationships between different targets. In other words, operations from global-dependency part implicitly depict the dependency between an interaction and the whole image, while operations from instance-dependency part compute the dependency between an interaction and the related instance.

In the detail implementation, we embed feature \emph{$\theta$},\emph{$\varphi$}, and \emph{g} onto a 512-dimensional space firstly. Then, we measure the similarity of \emph{$\theta$} and \emph{$\varphi$} in the embedding space using matrix multiplication and obtain the attentional weight by applying a softmax function, which corresponds to  \emph{$\frac{1}{C(x)} f(x_i,x_j)$} in Equation 1. Finally, we apply the attentional weight on feature \emph{g} by a multiplication operation. Here {$W_{\{\theta,\varphi,g\}}$} are the learnable parameters.
\begin{equation}
F(\theta,\varphi, g)=softmax(\theta^TW^T_\theta*W_\varphi\varphi)*gW_g
\label{equ5}
\end{equation}

\subsection{GID-Net}

\begin{figure}
	\centering
		\includegraphics[scale=.3]{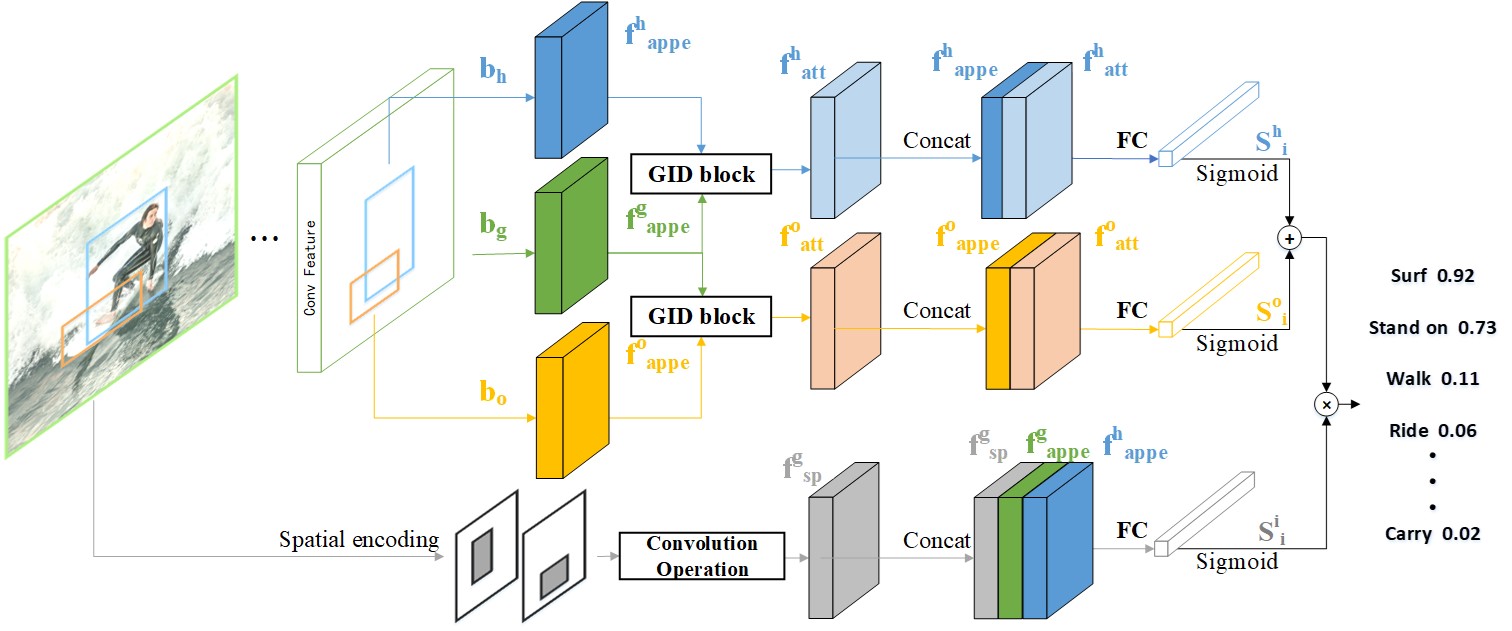}
	\caption{ An overview of our proposed GID-Net. The framework consists of three branches, which are a human branch, an object branch and an interaction branch. Given an image and the detected bounding boxes from Faster R-CNN, we predict every potential HOI among all $\langle$ person, object $\rangle$ pairs.}
	\label{FIG:3}
\end{figure}

HOI detection is regarded as not only a cognition problem, but also an inference problem which is based on the semantic information and fine-grained details from a scene. To detect a HOI, we need to accurately localize the human and the associated object, as well as identify the interaction between them. In our implementation, Faster R-CNN \cite{ren2017faster} has been adopted to detect all instances from the image. We denote the bounding boxes of the human and object by b$_h$ and b$_o$, respectively. We also express the box of the whole image as b$_g$. With the input image and the detected bounding boxes, we then generate proposals of $\langle$ b$_h$,b$_o$ $\rangle$ pairs and identify potential HOIs with the $\langle$ person, verb, object $\rangle$ triplets. For each human-object pair $\langle$ b$_h$,b$_o$ $\rangle$, we predict the score S$_i^{h,o}$ for each interaction \emph{$i\epsilon\{1,\cdots,I\}$}, where I denotes the total number of potential interactions. Figure~\ref{FIG:3} exhibits the overview of our proposed GID-Net. Given a $\langle$ b$_h$,b$_o$ $\rangle$ proposal, we detect the HOIs using a multi-stream network similar to InteractNet \cite{gkioxari2018detecting}, where different branches contribute to the detection with different features. Our proposed framework consists of a human branch, an object branch and an interaction branch.

\subsubsection{Human/Object branch}

 In human and object branches, the instance-level appearance feature f$_{appe}^h$ for human and f$_{appe}^o$ for object are extracted from the image. Then, reasoned feature f$_{att}^h$ and f$_{att}^o$ obtained from the GID block are concatenated to f$_{appe}^h$ and f$_{appe}^o$, respectively.  Differing from the HO-RCNN \cite{chao2018learning} or the iCAN \cite{gao2018ican:}, our human and object branches take image features from both Res\_4 and Res\_5 as inputs to encode long-range dependency. With this design, GID-Net is able to utilize the rich representation of features from different layers. Also, before determining the feature fusion mode as concatenation operations, we have tried different ways of aggregating reasoned features with appearance features (detailed in section 4.3). With the feature vectors, two fully connected layers followed by a softmax operation are employed to predict the confidence score related to the interactions, namely S$_i^h$ and S$_i^o$ respectively.

\subsubsection{Interaction branch}

 Although features from human and object branches contain strong semantic information for recognizing interactions, they are not sufficient for inferring spatial relationships between them. We adopt the two-channel binary image in \cite{chao2018learning} to encode the spatial relationship between the human and object, in which both the human box and object box are employed as the reference boxes to construct a binary image with two channels. The first channel has value \emph{1} at pixels enclosed by the human box and value \emph{0} elsewhere, while the second channel has value \emph{1} at pixels enclosed by the object box and value \emph{0} elsewhere. Several convolutional layers are adopted to extract the feature from the two-channel image, obtaining the spatial feature map f$_{sp}^g$.  After that, appearance features f$_{appe}^g$ from the global image and f$_{appe}^h$ from the human box are concatenated to f$_{sp}^g$ to predict the confidence score S$_i^i$. Our motivation of the concatenation above is that the global context and human appearance are both important for inferring an interaction and eliminating wrong interactions with similar spatial layouts. Our experiments show that global context are actually more helpful than human appearance to improve performance. It's worth mentioning that concatenating f$_{appe}^o$ from the object box with above features is not effective according to our experiments, since appearance of an object is usually constant in different kind of interactions. Also, some interactions do not involve any objects, e.g., $\langle$ person, walk $\rangle$.

\subsection{Multi-label classification}

As a person is likely to concurrently perform different interactions with one or multiple objects, HOI detection is a multi-label classification problem where each prediction of an interaction is independent and not mutually exclusive. Given a $\langle$ human, object $\rangle$ sample, we apply a binary sigmoid classifier for each interaction category \emph{$i\epsilon\{1,\cdots,I\}$}, and minimize the cross-entropy loss between the interaction score S$_i^h$, S$_i^o$ or S$_i^i$ and the ground-truth label for each interaction category. 

Taking human branch in our framework as an example, with a given image, let  \emph{$X^h  = [x_1^h,x_2^h,...,x_N^h]$} be predictions of \emph{N} $\langle$ human, object $\rangle$ samples, where \emph{$x_n^h\epsilon\Re^d$} is the feature vector of the \emph{n-th} sample. \emph{$Z^h  = [z_1^h,z_2^h,...,z_N^h]$} is the label indicative matrix for \emph{N} samples, where  \emph{$z_n^h\epsilon\Re^I$} is the label vector of the \emph{n-th} sample. The loss L$_h$ is calculated as follow:
\begin{equation}
L_h=\frac{1}{N}\sum_{n=1}^{N}avg(max(x_n^h,0)-x_n^h*z_n^h+log(1+exp(-abs(x_n^h ))))
\label{equ1}
\end{equation}
All three branches are trained jointly, where the overall loss is the sum of three losses for each interaction category: (1) the classification loss  L$_h$ from the human branch, (2) the classification loss L$_o$ from the object branch and (3) the classification loss  L$_i$ from the interaction branch.
\begin{equation}
L_{overall}=\sum_{branch\epsilon\{h,o,i\}}L_{branch}
\label{equ2}
\end{equation}
With the score S$_i^h$, S$_i^o$ and S$_i^i$ obtained from the human branch, object branch and interaction branch respectively, the final score S$_i^{h,o}$ is produced by a fusion function, which can be inferred as follow:
\begin{equation}
S_i^{h,o}=(S_i^h+ S_i^o) * S_i^i
\label{equ3}
\end{equation}
While some of the interactions do not involve any objects, we use S$_i^h$ * S$_i^i$ only to indicate the final score.

\section{Experiments and Evaluations}

\subsection{Experimental Setup}

\paragraph{ Datasets and evaluation metrics} We evaluate our method and compare it with the state-of-the-arts on two large-scale benchmarks, including Verbs in COCO (V-COCO) \cite{gupta2015visual} and Humans Interacting with Common Objects (HICO-DET) datasets \cite{chao2018learning}.

V-COCO \cite{gupta2015visual} includes 10,346 images, which is a subset of MS COCO dataset \cite{lin2014microsoft}. It contains 16,199 human instances in total and provides 26 common HOI annotations. Each person in V-COCO is annotated with a binary label for each interaction, indicating whether the person is performing the certain interaction. Thus, each person can perform multiple interactions at the same time. 

The newly released HICO-DET \cite{chao2018learning} contains about 48k images and 600 HOI categories over 80 object categories, which provides more than 150K annotated instances of $\langle$ person, object $\rangle$ pairs. There are three different HOI category sets in HICO-DET, which are: (a) all 600 HOI categories (Full), (b) 138 HOI categories with less than 10 training instances (Rare), and (c) 462 HOI categories with 10 or more training instances (Non-Rare). There also exist two different evaluation settings: (a) Default setting: all images both containing and not containing the target object category are evaluated. (b) Known Object setting: only images containing the target object category are evaluated. 

A detected $\langle$ person, verb, object $\rangle$ triplet is considered as a true positive if (1) it has the correct interaction label and (2) both the predicted human and object bounding boxes have IoU of 0.5 or higher with the ground-truth boxes. For evaluating the performance, we use role mean average precision (role mAP) \cite{gupta2015visual} on both V-COCO and HICO-DET. 

\paragraph{Implementation details} Following the protocol in \cite{gao2018ican:}, we generate human and object bounding boxes from images using the ResNet-50 version of Faster R-CNN \cite{ren2017faster}. Human boxes with scores higher than 0.8 and object boxes with scores higher than 0.4 are kept for detecting HOIs. We train our model in the end-to-end manner with Stochastic Gradient Descent (SGD), with a learning rate of 0.001, a weight decay of 0.0001, and a momentum of 0.9. We train our network for 300K and 1800K iterations on V-COCO and HICO-DET, respectively. Our program is implemented by Tensorflow on a GPU of GeForce GTX TITAN X.

\subsection{Overall Performance}

\paragraph{Performance on V-COCO} We compare our approach with several state-of-the-arts in this subsection. Results in terms of AP$_{role}$ are shown in Table~\ref{table1}. It can be seen that our proposed GID-Net has an AP$_{role}$(\%) of 45.4, achieving the best performance among all methods. Compared with the InteractNet \cite{gkioxari2018detecting} with also three branches, we achieve an absolute gain of 5.4 points, which is a relative improvement of 14\%. To have an apples-to-apples comparison with the existing best-performing model iCAN \cite{gao2018ican:}, we retrained it on V-COCO with the GPU of GeForce GTX TITAN X. We achieve an absolute gain of 1.2 point over the retrained iCAN, which quantitatively shows the efficacy of our approach.

\begin{table*}[htbp]
\centering
\caption{ Performance comparison with state-of-the-arts on V-COCO.}
\label{table1}
\begin{tabular}{llc}
\hline
Method & Backbone Network & AP$_{role}$(\%)  \\ \hline
Gupta et al. \cite{gupta2015visual} & ResNet-50-FPN & 31.8  \\
InteractNet \cite{gkioxari2018detecting} & ResNet-50-FPN & 40.0  \\
iHOI \cite{xu2018interact} & ResNet-50 & 40.4  \\
BAR-CNN \cite{kolesnikov2018detecting} & Inception-ResNet & 41.1  \\
GPNN \cite{qi2018learning} & Deformable ConvNet & 44.0  \\
iCAN retrained \cite{gao2018ican:} & ResNet-50 & \underline{44.2}  \\
\textbf{GID-Net (ours)} & ResNet-50 & \textbf{45.4}  \\  \hline
\end{tabular}
\end{table*}

\paragraph{Performance on HICO-DET} Table~\ref{table2} shows the comparisons of GID-Net with state-of-the-arts on HICO-DET. We report the quantitative evaluation of full, rare, and non-rare interactions with two different settings: Default and Known Object. It is happy to see that our method outperforms InteractNet with the same backbone network by an average of 5.1 points under the Default setting, amounting to a relative improvement of 54.9\%. Besides, the detection results of our method are higher than the second best under all evaluation settings. These results demonstrate that our GID-Net is more competitive than others in HOI detection.

\begin{table*}[htbp]
\centering
\caption{Results on HICO-DET test set. Default: all images. Known Object: only images containing the target object category. Full: all 600 HOI categories. Rare: 138 HOI categories with less than 10 training instances. Non-Rare: 462 HOI categories with 10 or more training instances.}
\label{table2}
\resizebox{\textwidth}{!}{
\begin{tabular}{llcccccc}
\hline
\multirow{2}{*}{Method} & \multirow{2}{*}{Backbone Network} & \multicolumn{3}{c}{default} & \multicolumn{3}{c}{know object}  \\ \cline{3-8}
 & & \multicolumn{1}{c}{full} & \multicolumn{1}{c}{rare} & \multicolumn{1}{c}{non-rare} & \multicolumn{1}{c}{full} & \multicolumn{1}{c}{rare} & \multicolumn{1}{c}{non-rare} \\ \hline
Shen et al. \cite{shen2018scaling} & VGG-19 &6.46 &4.24 &7.12 & - & - & - \\
HO-RCNN \cite{chao2018learning} & CaffeNet & 7.81 & 5.37 & 8.54 & 10.41& 8.94 & 10.85 \\
InteractNet \cite{gkioxari2018detecting} &ResNet-50-FPN &9.94 &7.16 &10.77 & - & - & - \\
iHOI \cite{xu2018interact} &ResNet-50 & 9.97 & 7.11  &10.83 & - & - & - \\
GPNN \cite{qi2018learning} &Deformable ConvNet & 13.11 & 9.34 & 14.23 & - & - & - \\
iCAN \cite{gao2018ican:} & ResNet-50 & \underline{14.84} & \underline{10.45}& \underline{16.15} & \underline{16.26} &\underline{11.33} & \underline{17.73} \\
\textbf{GID-Net (ours)}  & ResNet-50 & \textbf{15.41} & \textbf{11.07} & \textbf{16.71} & \textbf{16.92} & \textbf{12.56} & \textbf{18.23}  \\  \hline
\end{tabular}}
\end{table*}

\paragraph{Computational Complexity}  When it comes to efficiency, our framework provides its advantages over the methods adding additional contextual inputs (e.g., estimated human pose), since the reasoned features obtained from GID block are automatically learned from convolutional features and jointly trained. GID block is efficient also because the process of getting features f$_{im}^g$, f$_{appe}^h$ and f$_{appe}^o$ are highly shared. While iCAN \cite{gao2018ican:} also employed ResNet-50 as backbone and generated an attention model without human pose, we compare our method with iCAN on detection speed to study the efficiency of our method. We test both of two methods on V-COCO and HICO-DET under the same conditions, with a GPU of GeForce GTX TITAN X. As shown in Table~\ref{table6}, the detection speed of our GID-Net is basically identical with that of the most competitive iCAN. However, compared with iCAN, our GID-Net brings an absolute gain of 1.2 points on V-COCO and an absolute gain of 0.57 point on HICO-DET (default full).

\begin{table*}[htbp]
\centering
\caption{Comparison of detection speed and performance with iCAN \cite{gao2018ican:}.}
\label{table6}
\resizebox{\textwidth}{!}{
\begin{tabular}{llccc}
\hline
Method & Backbone Network & Detection Speed(fps) & \makecell[tl]{AP$_{role}$(\%) on \\ V-COCO }& \makecell[tl]{AP$_{role}$(\%) on \\ HICO-DET (default full)}  \\ \hline
iCAN retrained \cite{gao2018ican:} & ResNet-50  & 0.634 &44.2 & 14.84 \\
\textbf{GID-Net (ours)} & ResNet-50 & 0.597 & 45.4 & 15.41 \\  \hline
\end{tabular}}
\end{table*}

\subsection{Ablation Studies}

We adopt several ablation studies in this subsection. VCOCO and HICO-DET serve as the primary testbeds on which we further analyze the individual effect of components in our method.

\paragraph{Effects of decomposing GID block} We perform an ablation study on GID block by breaking it into two parts. Table~\ref{table4} summarizes the detailed performance of GID block with different parts on V-COCO. There are 24 interaction categories have been evaluated, in which 'cut', 'eat', 'hit' involve two types of target objects (instrument and direct object). It can be seen that, GID-Net without adopting GID block achieves an AP$_{role}$(\%) of 43.7 which already outperforms the InteractNet. When we only adopt the global-dependency part or the instance-dependency part in GID block, the AP$_{role}$(\%) are 44.1 and 44.3 respectively, indicating the respective roles of these two parts. Specifically, we can draw a relatively conclusion from the table that global-dependency part is more adept in depicting interactions closely related to the environment, for example, $\langle$ person, drink with, glass $\rangle$ (in a restaurant), $\langle$ person, throw, frisbee $\rangle$ (on the grass) or $\langle$ person, surf, surfboard $\rangle$ (in the sea). While instance-dependency part performs better in detecting interactions closely related to the human pose, for example, $\langle$ person, kick, ball $\rangle$ or $\langle$ person, talk on, phone $\rangle$. It is because global-dependency part captures dependency throughout the image while instance-dependency part pays more attention to instances. The adjustment boosts the performance by 5.4 points compared with InteractNet while both global-dependency part and instance-dependency part are adopted. The same experiments are adopted on HICO-DET dataset and results are summarized in Table~\ref{table7}. According to the results, both of the global-dependency part and instance-dependency part bring performance gain in detecting HOIs. The GID block brings an absolute gain of 5.1 points on average of the default setting compared with InteractNet. 

\begin{table*}[htbp]
\centering
\caption{The impact of decomposing GID block into two parts on V-COCO dataset. We show results of 24 interaction categories in V-COCO to get subtle comparisons.}
\label{table4}
\resizebox{\textwidth}{!}{
\begin{tabular}{lccccc}
\hline
Categories & InteractNet & \makecell[tl]{GID-Net w/o \\ GID block} & \makecell[tl]{GID-Net w/ \\ Global-dependency part} & \makecell[tl]{GID-Net w/ \\ Instance-dependency part} & \makecell[tl]{GID-Net w/ \\ GID block \\ (two parts together)}   \\ \hline
carry & 33.1 & 37.5  & 33.7 & 36.1 & 35.0     \\ 
catch & 42.5 & 43.7  & 43.6 & 44.3 & 45.5     \\ 
drink & 33.8 & 31.3  & 28.1 & 22.5 & 24.8     \\ 
hold & 26.4 & 26.5  & 25.2 & 24.9 & 25.6     \\ 
jump & 45.1 & 51.0  & 51.6 & 51.9 & 53.6     \\ 
kick & 69.4 & 56.3  & 62.2 & 64.0 & 65.0     \\ 
lay & 21.0 & 23.3  & 22.8 & 22.7 & 23.1     \\ 
look & 20.2 & 10.7  & 16.1 & 16.8 & 17.5     \\ 
read & 23.9 & 35.4  & 25.9 & 26.5 & 30.3     \\ 
ride & 55.2 & 67.8  & 65.2 & 65.1 & 64.3     \\ 
sit & 19.9 & 28.4  & 26.6 & 27.3 & 29.3     \\ 
skateboard & 75.5 & 80.9  & 81.6 & 82.1 & 82.6     \\ 
ski & 36.5 & 39.8  & 40.8 & 39.6 & 41.9     \\ 
snowboard & 63.9 & 68.4  & 72.8 & 73.4 & 73.1     \\ 
surf & 65.7 & 71.1  & 79.2 & 78.7 & 79.5     \\ 
talk\_on\_phone & 31.8 & 52.1  & 50.1 & 52.1 & 49.4     \\ 
throw & 40.4 & 36.3  & 41.5 & 40.2 & 45.6     \\ 
work\_on\_computer & 57.3 & 61.8  & 59.5 & 61.3 & 62.9     \\  \hline
cut-instr & 36.4 & 34.0  & 35.1 & 35.5 & 36.1     \\ 
cut-obj & 23.0 & 32.1  & 37.3 & 37.1 & 37.1     \\ 
eat-instr & 2.0 & 6.0  & 6.9 & 6.7 & 7.3     \\ 
eat-obj & 32.4 & 39.5  & 38.6 & 38.9 & 40.4     \\ 
hit-instr & 43.3 & 71.7  & 74.6 & 74.7 & 74.5     \\ 
hit-obj & 62.3 & 43.4  & 38.6 & 40.8 & 44.4         \\  \hline
mAP & 40.0 & 43.7 & 44.1 & 44.3 & 45.4         \\ \hline
\end{tabular}}
\end{table*}

\begin{table*}[htbp]
\centering
\caption{The impact of decomposing GID block into two parts on HICO-DET dataset.}
\label{table7}
\resizebox{\textwidth}{!}{
\begin{tabular}{lccccc}
\hline
Evaluation Settings & InteractNet & \makecell[tl]{GID-Net w/o \\ GID block} & \makecell[tl]{GID-Net w/ \\ Global-dependency part} & \makecell[tl]{GID-Net w/ \\ Instance-dependency part} & \makecell[tl]{GID-Net w/ \\ GID block \\ (two parts together)}   \\ \hline
default full         & 9.94 & 14.03  & 14.60 & 15.08 & 15.41     \\ 
default rare         & 7.16 & 9.23  & 9.94 & 10.90 & 11.07     \\ 
default non-rare     & 10.77& 15.12  & 15.99 & 16.33 & 16.71     \\ 
know object full     & -    & 15.38  & 16.23 & 16.58 & 16.92     \\ 
know object rare     & -    & 10.84  & 11.60 & 12.19 & 12.56     \\ 
know object non-rare & -    & 17.07  & 17.61 & 17.89 & 18.23     \\  \hline
\end{tabular}}
\end{table*}

\begin{figure}
	\centering
		\includegraphics[scale=.3]{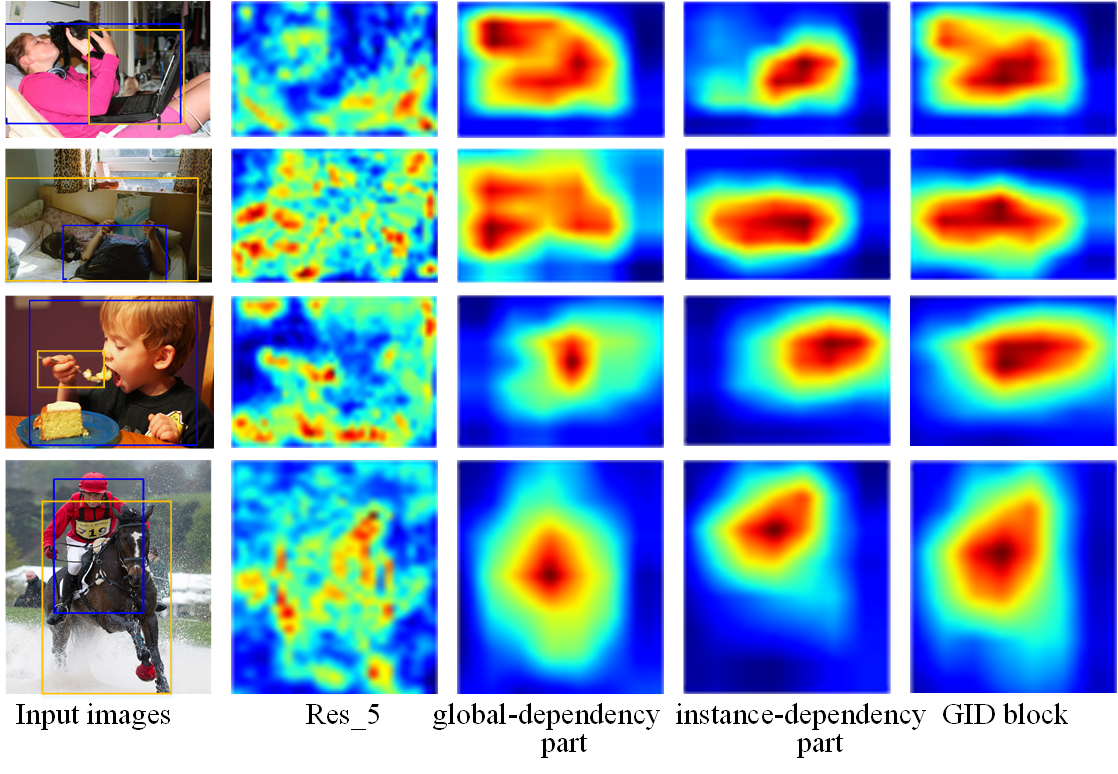}
	\caption{Visualization of our reasoning mechanism. The first column shows some randomly selected HOI examples with bounding boxes (i.e., $\langle$ person, work on, computer $\rangle$ , $\langle$ person, lie, bed $\rangle$, $\langle$ person, eat with, fork $\rangle$, $\langle$ person, ride, horse $\rangle$). The rest of columns visualize feature responses from Res\_5, global-dependency part, instance-dependency part and the whole GID block, respectively. In each subimage, the brightness of the pixel indicates how much the feature been noticed.}
	\label{FIG:8}
\end{figure}

{\color{red}Figure~\ref{FIG:8} visualizes the effects of our GID block. The first column shows original images with bounding boxes. Taking the human branch in our framework as an example, the rest of columns visualize feature responses from Res\_5 (the fifth residual block of ResNet-50), the global-dependency part of GID block, the instance-dependency part of GID block and the integrated GID block, respectively. Intuitively, two stages of the reasoning mechanism focus on learning different semantics, where the global-dependency part learns salient semantics throughout the image, and the instance-dependency part, on the other hand, mostly focuses on the informative semantics roughly correspond to the instances with on-going action. By adopting the strategy of multi-stage and progressive reasoning, our integrated GID block contains semantic-specific reasoning both in global-level and instance-level.}

\paragraph{Effects of feature fusion} GID block is a key component in our method. Table~\ref{table4} and Table~\ref{table7} have already showed that incorporating reasoned features from GID block can help to improve HOI detection. As shown in Table~\ref{table5}, we also investigate several different ways of incorporating reasoned features from the GID block with appearance features from the backbone network. Compared with addition operation and multiplication operation, concatenation is the most effective way to boost the performance, since it furthest maintains the diversity of features.  

\begin{table*}[htbp]
\centering
\caption{Comparisons of different ways of aggregating reasoned features from the GID block with appearance features from the backbone network.}
\label{table5}
\resizebox{\textwidth}{!}{
\begin{tabular}{lccccccc}
\hline
\multirow{2}{*}{Feature fusion operation} & \multirow{2}{*}{AP$_{role}$(\%) on V-COCO} & \multicolumn{3}{c}{HICO-DET default} & \multicolumn{3}{c}{HICO-DET know object}  \\ \cline{3-8}
 & & \multicolumn{1}{c}{full} & \multicolumn{1}{c}{rare} & \multicolumn{1}{c}{non-rare} & \multicolumn{1}{c}{full} & \multicolumn{1}{c}{rare} & \multicolumn{1}{c}{non-rare} \\ \hline
Addition        & 44.7 & 14.61  & 9.52 & 16.13 & 15.96 & 10.70 & 17.53 \\
Multiplication  & 44.5  & 14.34 & 9.23 & 15.92 & 15.77 & 10.43 & 17.46 \\
Concatenation   & 45.4  &15.41  & 11.07 & 16.71 & 16.92 & 12.56 & 18.23 \\  \hline
\end{tabular}}
\end{table*}

\paragraph{Effects of adopting different branches} We evaluate variants of our method when removing human branch or object branch in this subsection. We set the same parameters for the algorithm with previous sections but control the usage of different branches. As shown in Table~\ref{table3}, taking results on V-COCO as examples, removing the human branch and object branch from GID-Net reduces AP$_{role}$(\%) by 4.9 and 0.4 points, respectively. That is to say, while adopting just two branches for HOI detection, human branch shows greater strength than object branch, which can be interpreted as humans contribute more information than objects in an interaction. Thus, human-centric understanding has significant demand in practice. Aggregating all three branches yields the best AP$_{role}$(\%) of 45.4. 

\begin{table*}[htbp]
\centering
\caption{The impact of adopting different branches.}
\label{table3}
\begin{tabular}{lcccc}
\hline
Branch              &  \multicolumn{4}{c}{Usage}  \\ \hline
interaction branch  & \multicolumn{1}{c}{$\surd$} & \multicolumn{1}{c}{$\surd$} & \multicolumn{1}{c}{$\surd$} & \multicolumn{1}{c}{$\surd$}   \\
object branch       & \multicolumn{1}{c}{} & \multicolumn{1}{c}{$\surd$} & \multicolumn{1}{c}{} & \multicolumn{1}{c}{$\surd$}    \\
human branch        &\multicolumn{1}{c}{} & \multicolumn{1}{c}{} & \multicolumn{1}{c}{$\surd$} & \multicolumn{1}{c}{$\surd$}    \\   \hline
AP$_{role}$(\%) on V-COCO     & 40.5  & 41.8 & 45.0 & 45.4     \\  \hline
HICO-DET default full         & 13.34  & 14.33 & 15.16 & 15.41     \\ 
HICO-DET default rare         & 9.86  & 9.23 & 10.48 & 11.07     \\ 
HICO-DET default non-rare     & 14.37  & 15.85 & 16.52 & 16.71     \\ 
HICO-DET know object full     & 14.67  & 15.88 & 16.63 & 16.92     \\ 
HICO-DET know object rare     & 11.26  & 11.04 & 12.11 & 12.56     \\ 
HICO-DET know object non-rare & 15.69  & 17.32 & 17.94 & 18.23     \\  \hline
\end{tabular}
\end{table*}

\subsection{Error Analysis}

\begin{figure}
	\centering
		\includegraphics[scale=.53]{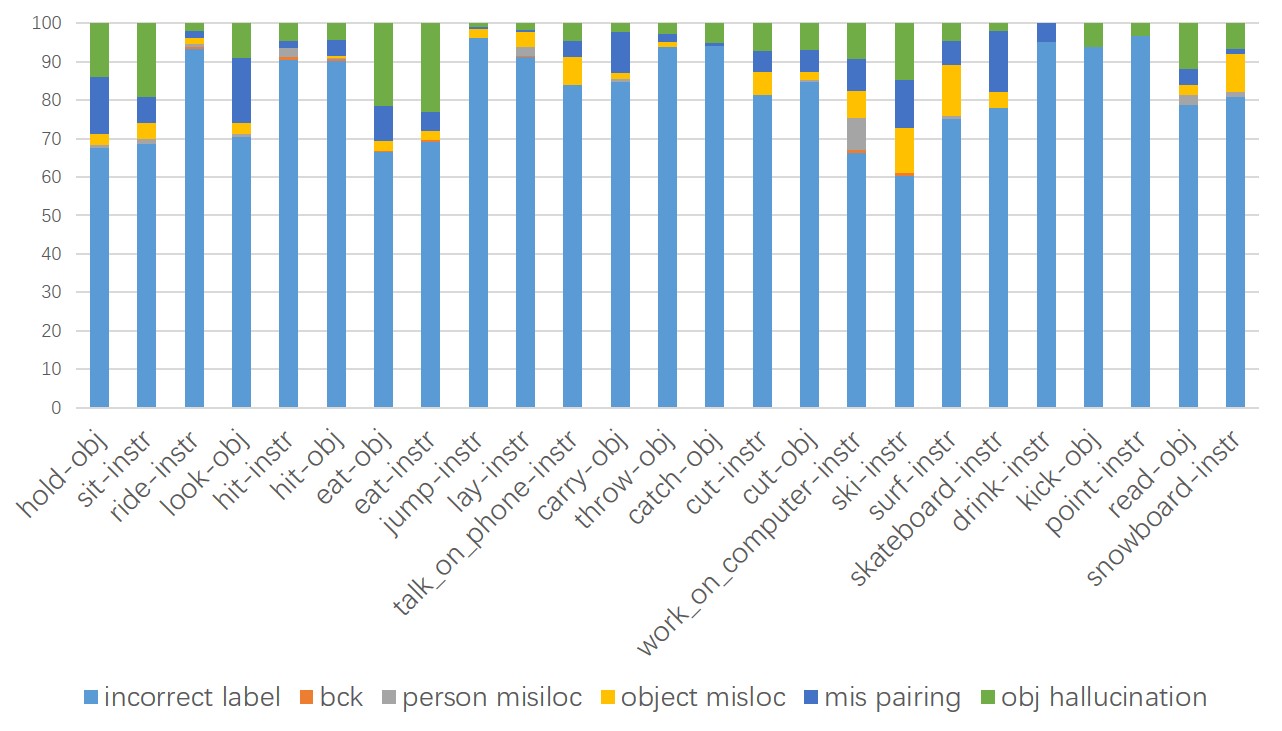}
	\caption{Distribution of the incorrect HOI detections for each interaction category on V-COCO. We diagnose the false positive detections in six types, which are incorrect label, bck, person misloc, object misloc, mis pairing and obj hallucination.}
	\label{FIG:4}
\end{figure}

\begin{figure}
	\centering
		\includegraphics[scale=.48]{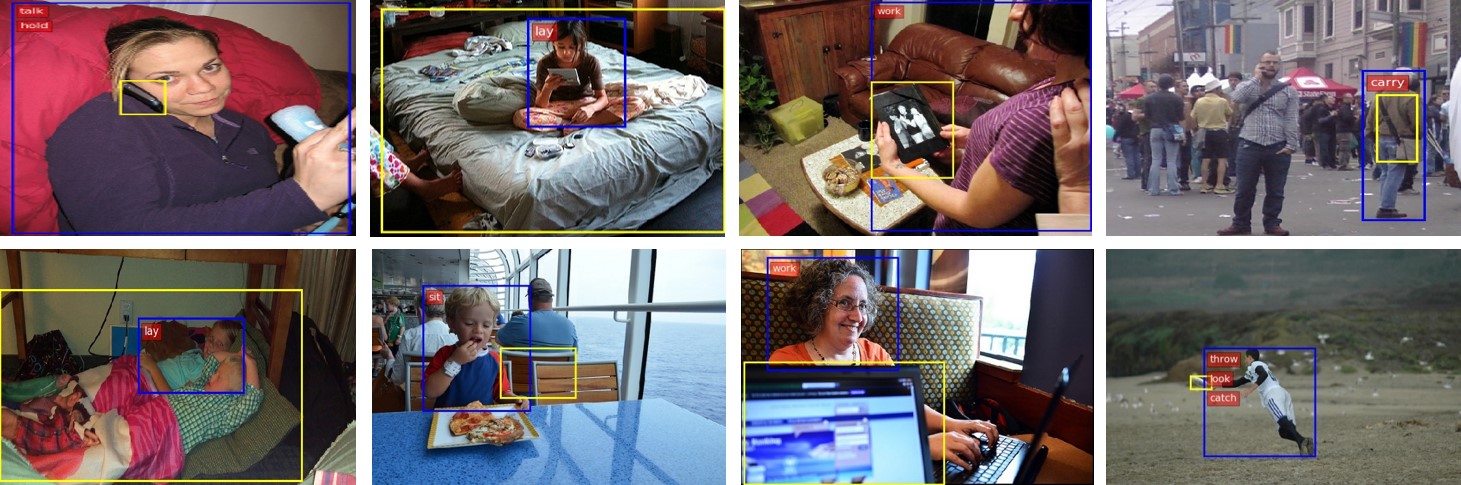}
	\caption{Visualizations of incorrect detections. The subplots show examples of incorrect label (first two), hallucination (third), miss-localization of both object (fourth) and human (fifth), Miss-pairing (sixth and seventh) and Semantic ambiguity (eighth) according to priority.}
	\label{FIG:7}
\end{figure}

To further examine the weakness of our method, we diagnose the detection results on V-COCO to produce a detailed error analysis of the following types \cite{gupta2015visual}: incorrect label, bck, person misloc, object misloc, mis pairing and obj hallucination. Figure~\ref{FIG:4} shows the distribution of incorrect detections from each interaction category. Meanwhile, some corresponding examples of incorrect detections are visualized in Figure~\ref{FIG:7}. Our method makes mistakes in detection mostly because of incorrect label, object miss-localization, object hallucination and miss-pairing. 

As shown, the most dominant error is incorrect label of interactions. Some of them are caused by the algorithm focusing too much on objects. For example, the algorithm is more likely to predict the interaction "hold" when it observes a phone and predict the interaction "lay" when it observes a bed (shown in the first two subplots in Figure~\ref{FIG:7}). This phenomenon partly illustrates that our exploited long-range dependency and reasoned features for HOI detection can be further developed to infer some kind of interactions.

Another common error is related to falsely recognized objects and humans, both by hallucination and miss-localization. The object in the third subplot of Figure~\ref{FIG:7} is incorrectly detected as a computer, leading to the algorithm predicts a wrong HOI label of "work". The fourth and fifth subplots in Figure~\ref{FIG:7} show examples of miss-localization of object and human, respectively. These errors could be potentially reduced by adopting more instance proposals from the object detector, but the number of $\langle$ person, object $\rangle$ pairs will increase quadratically, making the evaluation of all proposals infeasible.

Miss-pairing of HOI is also frequent in some scene. The sixth and seventh subplots in Figure~\ref{FIG:7} show some miss-pairing detections, in which objects are detected as associating with wrong person. This phenomenon indicates that capturing and enhancing the instance dependency is still an interesting open problem for future research.

Sometimes the interaction itself is semantically ambiguous. For example, it is extremely difficult to distinguish the interaction between "throw" and "catch" in the eighth subplot of Figure~\ref{FIG:7}. As a result, our method predicts both two labels with high confidence.

\subsection{Quantitive Examples}

\begin{figure}
	\centering
		\includegraphics[scale=.39]{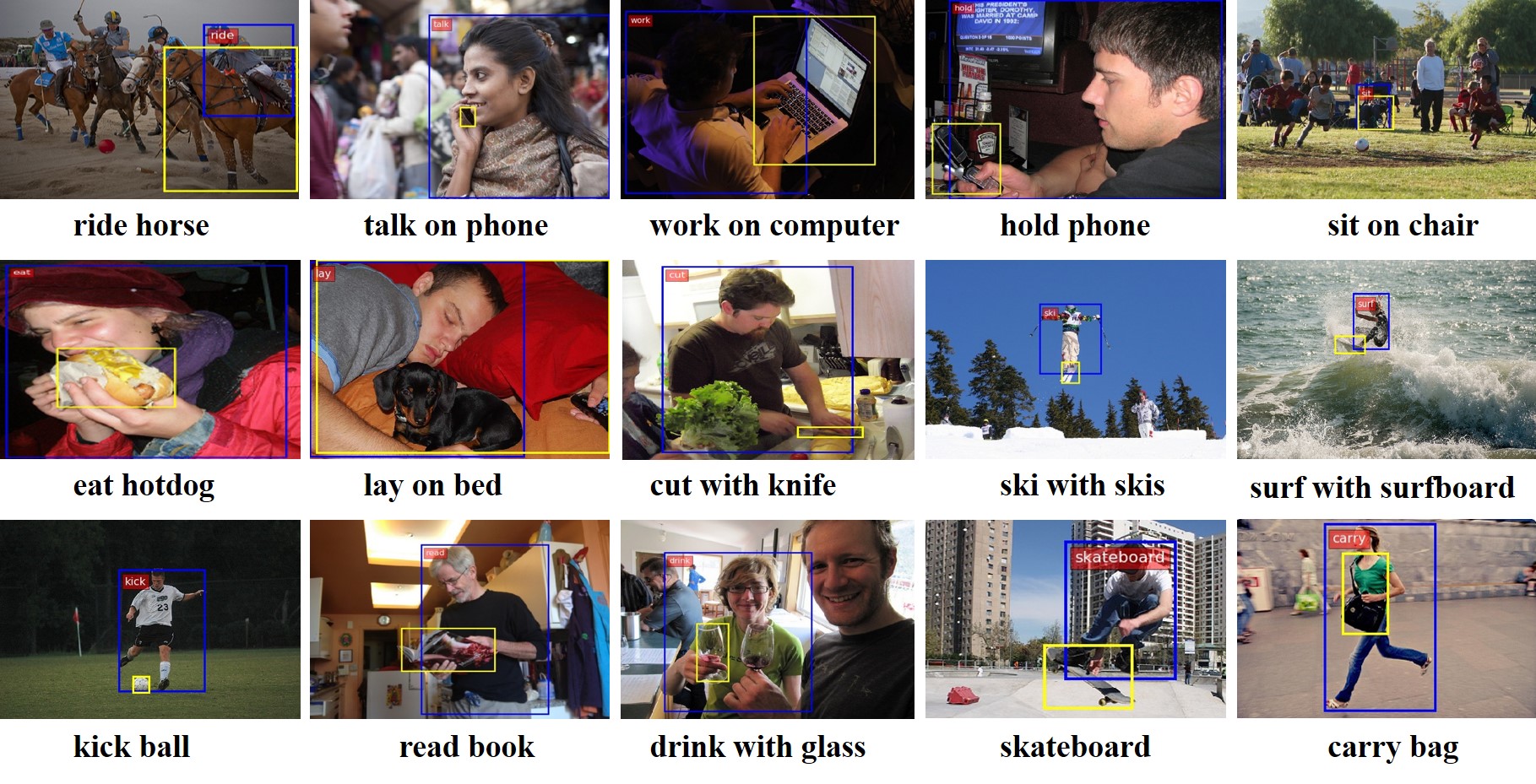}
	\caption{Detection results on V-COCO test set. Each subplot shows one detected $\langle$ person, verb, object $\rangle$ triplet. Blue boxes indicate the detected person and yellow boxes indicate the detected objects. The text below each subplot indicates the detected $\langle$ verb, object $\rangle$ tuple or just the verb for conciseness.}
	\label{FIG:5}
\end{figure}

\begin{figure}
	\centering
		\includegraphics[scale=.32]{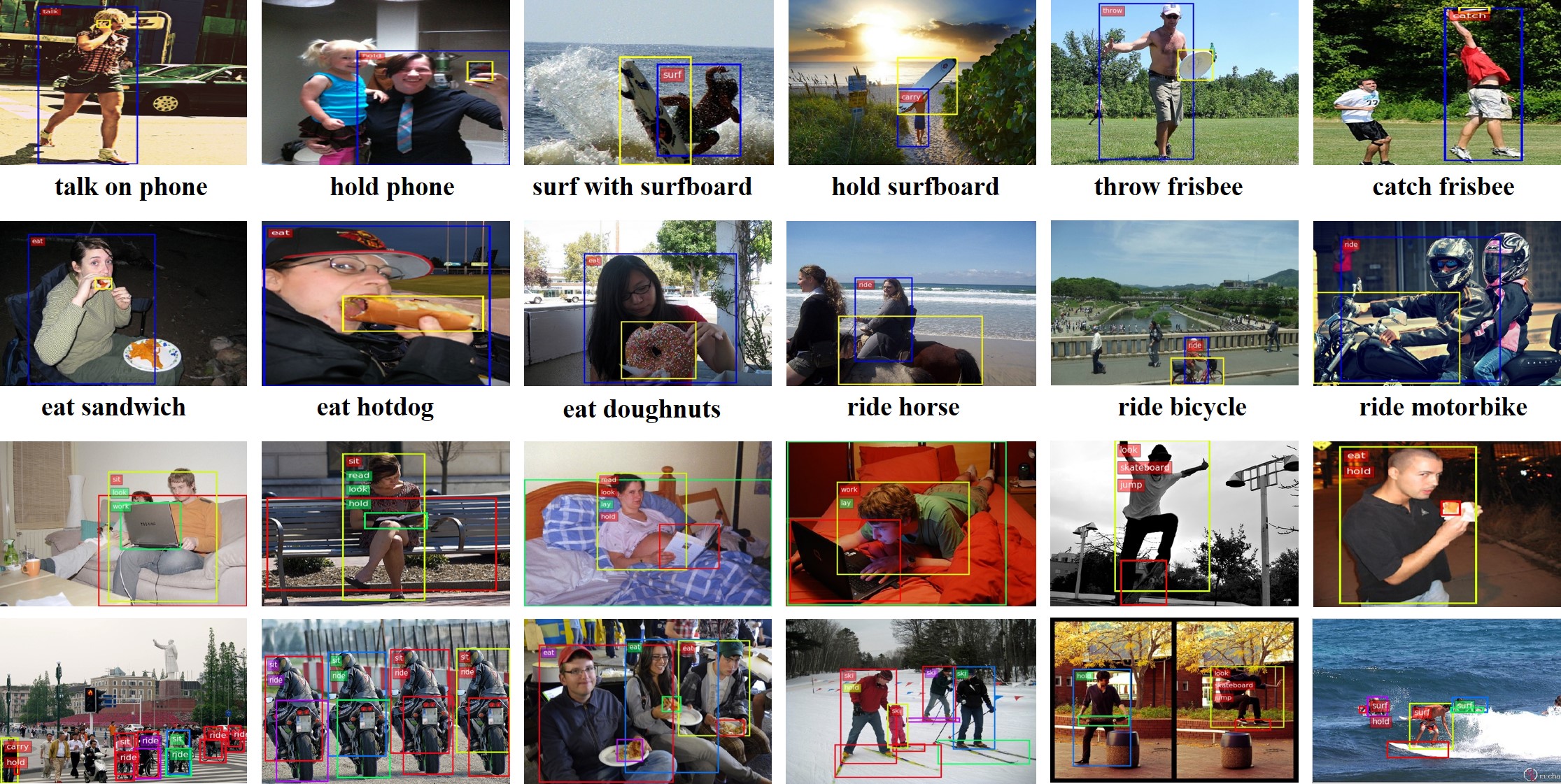}
	\caption{Detection results on test images. Our algorithm can detect various forms of HOIs in different scenes. First row: different interactions with the same objects. Second row: same interactions with different objects. Third row: one person takes interactions with various objects. Forth row: multiple people take interactions with various objects concurrently.}
	\label{FIG:6}
\end{figure}

For visualization purpose, several examples of detection are given in  Figure~\ref{FIG:5} and  Figure~\ref{FIG:6}. Each subplot in Figure~\ref{FIG:5} illustrates only one detected interaction for easy observation, including the location of detected person and object, as well as the interaction between the above two instances. We highlight the detected human and object with blue and yellow bounding boxes, respectively. Results from  Figure~\ref{FIG:5} indicate that our method is able to predict HOIs in a wide variety of situations.

Unlike object detection tasks that one instance has only one ground-truth label, HOI detection tasks may contain an instance with different labels in different situations. In Figure~\ref{FIG:6}, the first row shows that our algorithm is capable of predicting different interactions with the same objects (e.g., $\langle$ person, talk on, phone $\rangle$, $\langle$ person, hold, phone $\rangle$). The second row shows the results of predicting same interactions with different objects (e.g., $\langle$ person, ride, horse $\rangle$, $\langle$ person, ride, bicycle $\rangle$). The third row presents examples of detecting one person taking multiple interactions with surrounding objects. Moreover, as shown in the fourth row, our method is able to detect multiple people taking different interactions with various objects concurrently in an image. According to the results from Figure~\ref{FIG:6}, our proposed method can adapt to complex environments to provide high quality HOI detections.

\section{Conclusion}

In this paper, to remedy the limited capability of convolution operations in depicting visual interactions between humans and objects, we introduced a two-stage reasoning mechanism with a progressive process for HOI detection, namely GID block. Conforming to the human perception, GID block not only learns appearance features from instances, but also captures both global and instance dependency of pixels from the scene. Without employing any additional inputs (e.g., estimated human pose) or using any untrainable attention tricks, this is a significant try to produce semantic-specific reasoning both in global-level and instance-level in a single trainable mechanism. On this basis, we construct GID-Net, a three-stream HOI detection framework consists of a human branch, an object branch and an interaction branch. Extensive experiments have been conducted to evaluate our GID-Net on two large public benchmarks, which are V-COCO and HICO-DET. Our GID-Net outperforms existing best-performing methods and achieves the state-of-the-art results on both of the above two datasets. At last, we conduct comprehensive ablation studies and error analyses in GID-Net and hope to enlighten the future research. According to our error analysis, the most dominant detection error from the algorithm is incorrect label. Some of these mistakes are caused by the algorithm focusing too much on objects. Thus, enhancing the dependency from humans is still an interesting open problem in future research.

\section*{Acknowledgement} This paper was partially supported by National Engineering Laboratory for Video Technology - Shenzhen Division, and Shenzhen Municipal Development and Reform Commission (Disciplinary Development Program for Data Science and Intelligent Computing). Special acknowledgements are given to AOTO-PKUSZ Joint Research Center of Artificial Intelligence on Scene Cognition technology Innovation for its support.

\bibliography{myReference}

\end{document}